\documentclass[conference]{IEEEtran}

\usepackage{cite}
\usepackage{amsmath,amssymb}
\usepackage{graphicx}
\usepackage{booktabs}
\usepackage{subcaption}
\usepackage{url}
\usepackage{flushend}
\usepackage[hidelinks]{hyperref}

\graphicspath{{figures/}}

\begin{document}

\title{Catalogue Photography as a Cold Start: Toward Deployable Rotary Milling Tool Recognition}

\author{%
\IEEEauthorblockN{Abilash Philip Madavath\textsuperscript{*}, Chandra Yuvesh Aubeeluck\textsuperscript{*}, Augustin Raju\textsuperscript{*}}
\IEEEauthorblockN{Nicolas Pyschny\textsuperscript{*}, Felix Hackelöer\textsuperscript{*}, Florian Zwanzig\textsuperscript{*}}
\IEEEauthorblockA{\textsuperscript{*}TH K{\"o}ln -- University of Applied Sciences, Cologne, Germany}
}
\maketitle

\begin{abstract}
Verifying that manufactured batches of rotary milling tools, also known as carbide burrs, conform to production order sheets remains a largely manual and error-prone quality assurance task. Automating this process with computer vision faces a critical cold-start constraint since no labelled imagery from the deployment environment is available, leaving manufacturer catalogue photography as the sole source of supervision. We investigate how far catalogue supervision can support an industrial recognition pipeline under domain shift, explicitly measuring the gap between catalogue separability and performance on held-out field photographs. Our findings reveal three key insights. First, off-the-shelf frozen feature extractors do not reliably separate the two task attributes, head shape and tooth profile, motivating targeted representation learning. Second, metric learning produces near-perfect unsupervised cluster discovery on catalogue images (adjusted Rand index 0.94--0.97), yet on field photographs under half of the accuracy gained from training survives. Third, the largest transfer gains do not come from model scale or representation complexity, but from simple changes that reduce domain sensitivity: converting images to grayscale (+0.22) and constraining retrieval against the known order sheet (+0.11). We therefore treat catalogue photography as a useful cold start rather than a deployment-ready training domain, and provide empirical baselines and an evaluation protocol for catalogue-to-field transfer in precision tool manufacturing.
\end{abstract}


%==============================================================
\section{Introduction}
%==============================================================

\begin{figure}[!t]
\centering
\begin{subfigure}[b]{0.31\columnwidth}
  \centering
  \includegraphics[height=24mm]{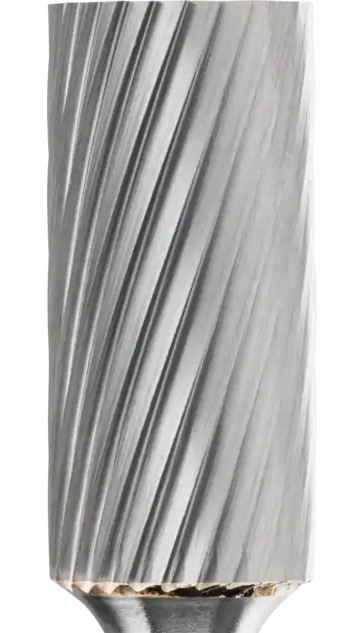}
  \caption{ZYA}
\end{subfigure}\hfill
\begin{subfigure}[b]{0.31\columnwidth}
  \centering
  \includegraphics[height=24mm]{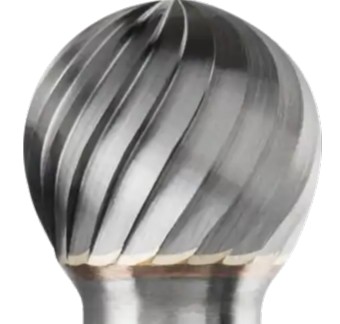}
  \caption{KUD}
\end{subfigure}\hfill
\begin{subfigure}[b]{0.31\columnwidth}
  \centering
  \includegraphics[height=24mm]{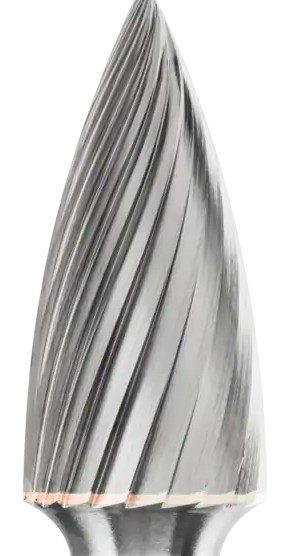}
  \caption{SPG}
\end{subfigure}
\caption{Three catalogue articles with the same cut (Z3) on different head shapes. The tooth pattern lines curve differently on each head, which is why one global appearance model mixes the two attributes. Images courtesy of August R{\"u}ggeberg GmbH \& Co.~KG (PFERD) \cite{pferd}.}
\label{fig:tools}
\end{figure}

Visual inspection in manufacturing is still largely manual, and a recent survey shows why this reaches its limits: error rates on complex inspection tasks range from twenty to thirty percent, made worse by fatigue and lapses in attention, and fewer than six percent of the $196$ studies reviewed ask whether the correct parts are present in the first place \cite{huetten2024survey}. That last question is exactly ours. Rotary milling tools (\emph{Frässtifte}, also called carbide burrs) come in several hundred variants, and after production an operator has to confirm that the tools on a pallet match the article numbers on a scanned order sheet. Two tools can share a silhouette and differ only in tooth pitch or cut direction (Fig.~\ref{fig:tools}).

The obstacle is that no packaging line imagery of the tool exists before the system is installed, and collecting it would interrupt production. The only training dataset is manufacturer catalogue photography: studio-lit, colour-graded renderings that differ systematically from what a camera above a pallet will see, where specular reflections and line lighting dominate. Catalogue accuracy measured on its own is therefore misleading; the catalogue-to-field gap ($\Delta_{\text{acc}}$) is the number that matters \cite{zhu2023simtoreal,torralba2011unbiased}. We treat this as a \emph{cold start}, a supervised source with no exposure to the deployment domain, and ask how far it carries a recognition system.

Our approach describes each tool by two independent properties rather than as one of thousands of separate types, following work on compositional recognition, which addresses the case where classes are combinations of attributes and collecting data for every combination is infeasible \cite{naeem2021czsl}. For these tools the two properties are already standardised: head shape follows DIN~8032 and the cut follows DIN~8033 \cite{pferd}. What we add is a test of that idea against real photographs, and an honest account of what did and did not transfer.

This paper studies the feasibility of training on catalogue images alone, comparing models, feature representations and scoring strategies rather than proposing a single final pipeline. Camera hardware, packaging line integration and continual learning are out of scope. By \emph{catalogue-only} we mean that no image from the field set enters training, validation or model selection. We call the held-out real photographs the \emph{field set}, to keep them apart from the held-out \emph{catalogue split}.

%==============================================================
%==============================================================

%==============================================================
\section{Task, Data and Protocol}
\label{sec:formulation}
%==============================================================

We recognise a tool by two attributes: head shape $s\in\mathcal{S}$ and tooth profile $p\in\mathcal{P}$. Here $|\mathcal{S}|=9$ shapes (ZYA, KUD, SPG and six others) set by DIN~8032, and $|\mathcal{P}|=15$ cuts (pitches Z1--Z5 alongside material-specific geometries such as INOX or ALU) set by DIN~8033 \cite{pferd}.

Full article numbers also specify diameter and shank length, but those are scale-dependent and better resolved by camera calibration, a reference marker in the scene, or the order sheet itself. Working from scale-normalised head crops, our pipeline predicts only the pair $(s,p)$. That space is thin and very uneven: many combinations are never made, and those that exist range from fewer than ten images to over a hundred.

\subsection{Why two attributes rather than one label}
The same cut on a different head looks different (Fig.~\ref{fig:tools}). A model trained on profile labels \emph{across} head shapes has to learn to ignore this; a model trained on combined $(s,p)$ labels sees Z3-on-ZYA and Z3-on-KUD as unrelated classes and never has to. Whether this matters in practice is tested in Sec.~\ref{sec:clustering} and~\ref{sec:transfer}.

\subsection{The order sheet as a constraint}
\label{sec:constraint}
Matching each tool against the whole catalogue is an open-set problem over
hundreds of classes. But in our deployment setting each pallet carries a
single tool type, and the order sheet names that one expected article before
the pallet is imaged. The task is therefore not open-set search but
verification against a known target: does the observed tool's similarity to
the expected article clear a threshold, rather than which of many candidates
it best matches. We test this constrained-scoring approach against
unconstrained retrieval in Sec.~\ref{sec:factors}.

\subsection{Data}
The catalogue dataset has $770$ images over $9$ shape classes and $838$ over $15$ profile classes, scraped from manufacturer pages, cropped to the head and orientation-normalised; classes with fewer than five images were dropped. The two sets index the same photographs under different labels, giving $724$ images with both.

The field set is $45$ real photographs covering $7$ of $9$ shapes and $52$ covering $7$ of $15$ profiles. These pools are \emph{disjoint}, since no field image carries both labels, so no joint accuracy is measurable. Where a single field number is useful, we report the pooled top-1 rate over both pools, which is measured rather than assumed. A true joint accuracy would likely sit below it: on a real conveyor line, glare on specular carbide or a defocused crop would degrade shape and profile together.

\subsection{A leakage channel to watch for}
The catalogue lists the same physical tool under several surface treatments (uncoated, coated, specially finished). A random split puts near-duplicates on both sides. In our case that inflated catalogue accuracy to about $0.97$ while the field set sat near $0.51$. We now group images by parsed tool identity and check that no identity crosses the split. All catalogue numbers below are post-fix.

\subsection{Metrics}
\label{sec:metrics}
\emph{Recall at rank 1 (R@1)} asks whether an image's nearest neighbour in feature space carries the same label, that is, whether similar tools land close together, with no clustering step. \emph{Adjusted Rand index (ARI)} clusters the features and scores how well those clusters line up with the true classes: $1$ is a perfect match, $0$ is no better than random. R@1 measures local neighbourhoods, ARI the global shape of the space, and the two can disagree. \emph{Top-1 accuracy (acc)} is the usual recognition score. On the field set we report it \emph{pooled}: each of the $45$ shape photographs and $52$ profile photographs yields one right-or-wrong prediction on its own attribute, and pooled accuracy is the fraction correct over all $97$. It is a directly counted rate, not a claim that both attributes were correct on the same tool. Finally, field gap $\Delta_{\text{acc}} = \mathrm{acc}_{\text{catalogue}} - \mathrm{acc}_{\text{field}}$ measures how much of a configuration's catalogue score is rendering style rather than tool identity; lower is better, and ranking by catalogue accuracy alone rewards the wrong property.

\subsection{Statistical limits}
\label{sec:power}
With $45$ and $52$ field images, the 95\% bootstrap interval on a single accuracy is roughly $\pm 0.10\text{--}0.15$, and a paired McNemar test between the leading configurations is not significant. We therefore treat the top approaches as statistically equivalent and use the field set to detect domain shift, not to rank.

%==============================================================
\section{Four Stages}
\label{sec:validation}
%==============================================================

%--------------------------------------------------------------
\subsection{Stage 1: frozen extractors}
\label{sec:featurizers}
%--------------------------------------------------------------

We first asked whether generic pretrained backbones separate milling tool morphology out of the box. We ran $41$ feature extractors through six clustering algorithms ($k$-means, GMM, agglomerative with Ward linkage, spectral clustering, affinity propagation and HDBSCAN) on each attribute, covering self-supervised transformers \cite{oquab2024dinov2,simeoni2025dinov3}, supervised ImageNet CNNs and ViTs \cite{tan2019efficientnet,liu2022convnext}, and a geometric contour descriptor. Spectral clustering and affinity propagation are excluded from the reported figures: the former degrades on tightly separated clusters, the latter overshoots the cluster count by an order of magnitude. All ARI values below are the best over the remaining four.

\begin{table}[!ht]
\caption{Stage 1. Top five frozen extractors per attribute, by best ARI across four clustering algorithms.}
\label{tab:top5}
\centering
\footnotesize
\begin{tabular}{llcc}
\toprule
Featurizer & Best clusterer & ARI & R@1 \\
\midrule
\multicolumn{4}{l}{\textbf{Head shape}} \\
DINOv3 ViT-L/16     & GMM           & \textbf{0.34} & \textbf{0.89} \\
DINOv2 ViT-S/14     & GMM           & 0.29          & 0.63          \\
ResNet-18           & $k$-means     & 0.26          & 0.59          \\
DINOv2 ViT-B/14     & GMM           & 0.25          & 0.61          \\
Contour descriptor  & $k$-means     & 0.25          & 0.60          \\
\midrule
\multicolumn{4}{l}{\textbf{Tooth profile}} \\
EfficientNet-B4     & Agglomerative & \textbf{0.33} & \textbf{0.77} \\
ConvNeXt-L          & Agglomerative & 0.27          & 0.77          \\
Wide-ResNet-101     & Agglomerative & 0.27          & 0.76          \\
EfficientNet-B3     & Agglomerative & 0.27          & 0.75          \\
EfficientNet-B2     & Agglomerative & 0.26          & 0.76          \\
\bottomrule
\end{tabular}
\end{table}

The two lists in Table~\ref{tab:top5} share no entry. Self-supervised transformers win shape, led by DINOv3 ViT-L/16 (ARI $0.34$, R@1 $0.89$), with even a hand-built contour descriptor in the top five. Supervised CNNs win profile, led by EfficientNet-B4 (ARI $0.33$), where DINOv3 collapses to $0.10$. Scale brings no benefit: across EfficientNet B0--B7 and ConvNeXt tiny--large the scores vary erratically with no trend. HDBSCAN, the one algorithm not given the true class count, never wins a row. This is a sign that frozen feature spaces lack clear cluster boundaries, and it reverses in Stage 2. No single frozen representation handles both attributes, so specialisation has to come from training. The best frozen setup becomes baseline \textbf{A1}.

%--------------------------------------------------------------
\subsection{Stage 2: what training does to the features}
\label{sec:clustering}
%--------------------------------------------------------------

We then trained on catalogue imagery under two schemes. \textbf{A2} is a shared backbone with two classification heads. \textbf{A4} is two separate backbones trained with an additive angular margin loss \cite{deng2019arcface}, one for shape and one for profile. The clustering protocol is unchanged.

\begin{table}[!ht]
\caption{Catalogue separability before and after training.}
\label{tab:beforeafter}
\centering
\footnotesize
\begin{tabular}{lcccc}
\toprule
& \multicolumn{2}{c}{\textbf{Head shape}} & \multicolumn{2}{c}{\textbf{Tooth profile}} \\
\cmidrule(lr){2-3}\cmidrule(lr){4-5}
Representation & ARI & R@1 & ARI & R@1 \\
\midrule
Best frozen (Stage 1)$^{\ast}$ & 0.34 & 0.89 & 0.33 & 0.78 \\
A2: multi-task backbone        & 0.86 & \textbf{1.00} & 0.34 & 0.96 \\
A4: two-stream ArcFace         & \textbf{0.97} & \textbf{1.00} & \textbf{0.95} & \textbf{0.98} \\
\bottomrule
\end{tabular}
\\[2pt]
\raggedright\scriptsize $^{\ast}$Best value per column, so the model behind each
entry differs; the profile R@1 of $0.78$ comes from EfficientNet-B1, which does
not appear in Table~\ref{tab:top5} because that table ranks on ARI.
\end{table}

\begin{figure}[!ht]
\centering
\includegraphics[width=\columnwidth]{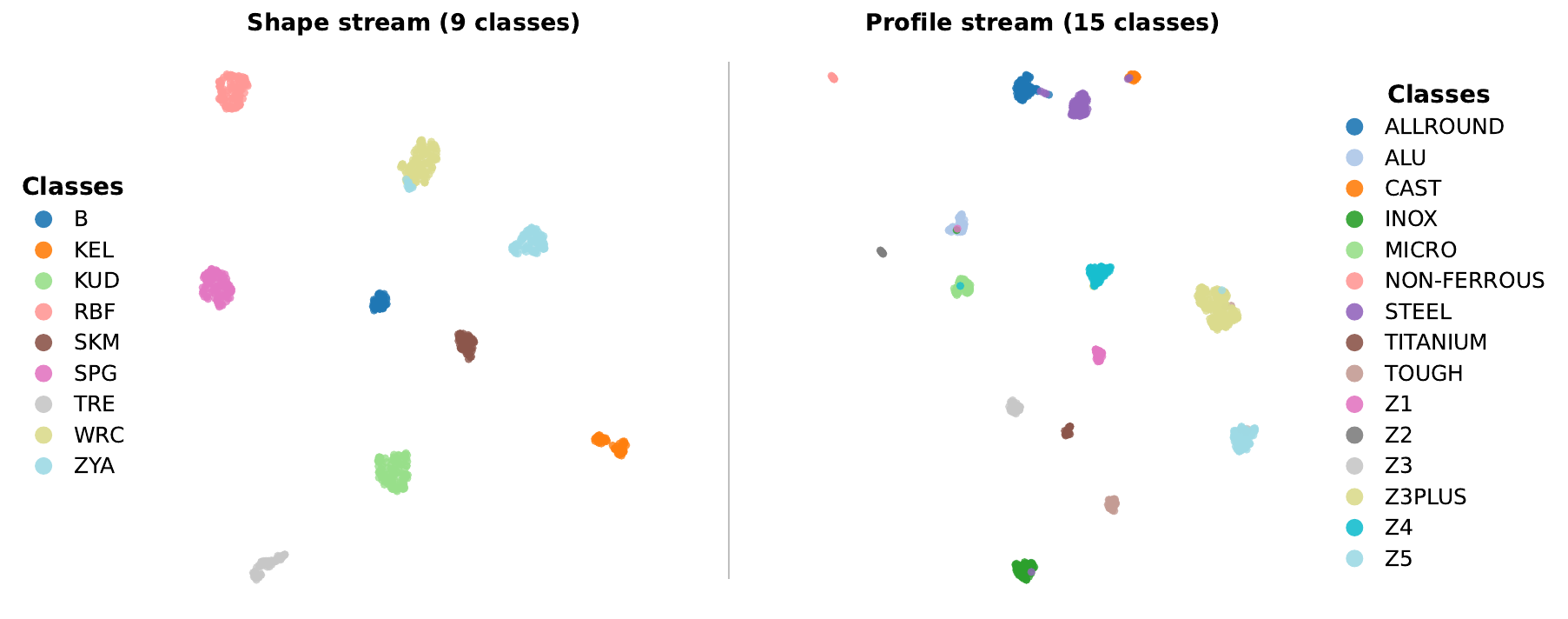}
\caption{UMAP projection of catalogue embeddings from the trained two-stream model (A4), coloured by ground-truth label. Each stream forms tight, well-separated groups for its own attribute.}
\label{fig:scatter}
\end{figure}

Training changes the picture completely (Table~\ref{tab:beforeafter}, Fig.~\ref{fig:scatter}). A4 reaches ARI $0.94\text{--}0.97$ under all four clustering algorithms at once, and HDBSCAN \cite{campello2013hdbscan}, given no class count, finds exactly $9$ shape clusters and $15$ profile clusters. That is a stronger check than a fixed-$k$ method hitting a target it was handed. The cross-check holds too: shape-trained features score ARI below $0.03$ against profile classes and the reverse, so each stream knows only its own attribute.

One caution on the metrics. A2 has excellent nearest-neighbour recall (R@1 $0.96\text{--}1.00$) yet its profile ARI stalls at $0.34$, matching the untrained baseline, and HDBSCAN breaks its profile space into $71\text{--}89$ small fragments rather than $15$ groups. Local retrieval can stay high while the global structure is broken up, which is why we report both.

%--------------------------------------------------------------
\subsection{Stage 3: what survives real photographs}
\label{sec:transfer}
%--------------------------------------------------------------

We compare three configurations on identical splits: \textbf{A1}, the best frozen features; \textbf{A3}, a single backbone trained with ArcFace \cite{deng2019arcface} on joint labels, the standard retrieval setup and our no-decomposition control; and \textbf{A4}. Six further approaches ran on the same harness; three fall below A1 in the field.

\begin{table}[!ht]
\caption{Top-1 accuracy on the catalogue split and the field set.}
\label{tab:leaderboard}
\centering
\footnotesize
\begin{tabular}{lccccc}
\toprule
& Cat. & \multicolumn{4}{c}{Field set} \\
\cmidrule(lr){2-2}\cmidrule(lr){3-6}
Configuration & Comb. & Shape & Profile & Pooled$^{\ast}$ & $\Delta_{\text{acc}}$ \\
\midrule
A1 (frozen)      & 0.371 & 0.756 & 0.413 & 0.300 & 0.071 \\
A3 (one stream)  & 0.857 & 0.863 & 0.593 & 0.522 & 0.335 \\
A4 (two streams) & \textbf{0.916} & \textbf{0.889} & \textbf{0.603} & \textbf{0.556} & 0.360 \\
\bottomrule
\end{tabular}
\\[2pt]
\raggedright\scriptsize $^{\ast}$Top-1 rate over both field pools combined ($97$ images), not a joint accuracy: no field image carries both labels.
\end{table}

Values are means over three seeds and two colour modes, with a seed spread of at most $\pm 0.05$ on the pooled field number. A4 leads nominally on both splits, but per Sec.~\ref{sec:power} it is not statistically separable from A3.

Training helps, but less than the catalogue suggests. Moving from A1 to A4 adds $0.545$ on the catalogue split and $0.256$ on real photographs: $47\%$ of the gain transfers, while $\Delta_{\text{acc}}$ grows five-fold.

The sharpest case is not in the table. An edge-map encoder trained with a supervised contrastive loss reaches ARI $0.85\text{--}0.91$ on catalogue profile, second only to A4, yet in the field it does worse than the untrained baseline. High catalogue separability can be an artefact of studio rendering rather than transferable geometry.

The case for A4 over A3 rests on specialisation, not accuracy. In the field A4's shape stream scores $0.91$ on shape and $0.22$ on profile, and the profile stream mirrors it ($0.22$ / $0.64$). That separation matters because order-sheet scoring needs roughly independent per-attribute costs, which one combined embedding cannot supply.

%--------------------------------------------------------------
\subsection{Stage 4: what actually moved the field number}
\label{sec:factors}
%--------------------------------------------------------------

Since neither the frozen features nor the label split gave a large transfer gain, we tested eight changes outside the architecture: colour mode, resolution, pooling, augmentation, feature adaptation, retrieval post-processing, backbone choice and the order-sheet constraint. Each was compared against the incumbent setup on the same images, keeping only effects above the measurement noise.

\begin{table}[!ht]
\caption{Effect of individual changes on the field set.}
\label{tab:ablation}
\centering
\footnotesize
\begin{tabular}{lcl}
\toprule
Change & Field gain & Outcome \\
\midrule
Grayscale (train and eval)   & $+0.22$ & kept \\
Order-sheet constraint       & $+0.11$ & kept$^{\dagger}$ \\
Covariance alignment (CORAL) & $-0.09$ & dropped (sample size too small) \\
Augmented gallery            & $-0.03$ & dropped (index clutter) \\
\bottomrule
\end{tabular}
\\[2pt]
\raggedright\scriptsize $^{\dagger}$Corrected 95\% CI $[+0.06,+0.17]$; measured
with synthetic candidate articles added to the sheet, since our real pallets
carry a single tool type and a one-candidate sheet gives no decision to score.
\end{table}

Only two survived (Table~\ref{tab:ablation}). Resolution and pooling produced variations indistinguishable from noise. Across seven alternative backbones only EfficientNet-B4 matched the incumbent within error, while DINOv2, SigLIP2, EVA02, ConvNeXtV2 and ResNet-50 all lost accuracy. The best post-processed frozen configuration reached an estimated combined accuracy (the product of its two attribute rates) of $0.48$, below A4's pooled $0.56$ even on the easier constrained task, so inference-time constraints do not replace representation learning.

Two caveats bound the order-sheet result. It was measured on frozen features only, and transferring it to the trained two-stream system is ongoing work. It was also measured with distractor articles padded onto the sheet: our pallets always carry one tool type, so the gain reflects constraining the search space rather than resolving competition between tools, and does not directly establish a deployment number for the single-target case. Neither surviving change touches the representation.

%==============================================================
\section{Discussion}
\label{sec:discussion}
%==============================================================

\textbf{Colour is the main shift.} Grayscale conversion provided the largest performance boost (+0.22). Catalogue colour grading functions as a misleading shortcut that the network learns during training, creating a chromatic domain shift that the factory line cannot replicate.

\textbf{Path to operational deployment.} A pooled field accuracy near $0.56$ cannot autonomously replace manual inspection, which demands reliability in excess of $99\%$. However, this figure reflects unconstrained open-set retrieval; incorporating the order-sheet constraint measured in Sec.~\ref{sec:factors} suggests headroom to lift operational accuracy, though that gain was measured under a synthetic multi-candidate condition and its size for our true single-target pallets remains to be confirmed. Beyond raw accuracy, the core utility of this catalogue cold start is operational: it delivers a day-one deployment baseline without prior imagery, specialized along standardized physical axes, while escalating low-confidence predictions to human operators. This human-in-the-loop framework converts routine inspection into an active-learning pipeline, where flagged edge cases generate the annotated in-domain imagery currently missing. We therefore treat $0.556$ not as an upper bound, but as a practical operational starting point.

\textbf{What did not work.} A cascade predicting shape first and routing to a shape-specific profile model did worse than the parallel two-stream design, despite removing the wrapping problem by construction: the per-shape data was too thin. CORAL failed because the field set is too small to estimate a covariance, and augmented-gallery expansion failed because we added views as separate index entries instead of averaging them. Both are implementation limits, not verdicts on the methods.

\textbf{Visual shortcuts in catalogue data.} A split meant to test invariance to photographic style clustered by tool shape instead, because product series and their photography setups are confounded with tool geometry in the catalogue. Even the best candidate separated styles only by proxy: style agreement tracked shape agreement almost one-to-one, while profile separation stayed poor. The feature space recovered the labels we wanted but not the nuisance we wanted removed. Anyone building a benchmark from a commercial catalogue should audit for this.

%==============================================================
\section{Limitations and Conclusion}
%==============================================================

The field set binds everything above. At $45$ and $52$ images it cannot separate the leading approaches, and until it covers the missing classes no deployment confidence interval would be honest. Expanding it comes before further model work. A paired field set, one tool photographed once and labelled on both attributes,
would turn the pooled rate reported here into a true joint accuracy and expose the correlated failures a pooled rate cannot show. The field photographs are also not packaging line images, so the gap we report is a lower bound.

Catalogue photography is a usable cold start. No frozen extractor separates the two attributes, so the decomposition must be trained in; once it is, four clustering algorithms agree the structure is there and rediscover the class counts unsupervised. But most of that lives in the catalogue: under half of training's gain survives real photographs, and an approach near the top on the catalogue sits below the untrained baseline in the field. The obstacle is domain shift, not model capacity, and the two changes that helped most, removing colour and using the order sheet, are not changes to the model at all.

\section*{Acknowledgment}
This work is supported through the InnoFaktur project by the European Regional Development Fund under grant EFRE-20500002. The authors thank August R{\"u}ggeberg GmbH \& Co.~KG (PFERD) for domain access and support.

\end{document}